\documentclass[journal]{IEEEtran}

\usepackage[noadjust]{cite}
\usepackage{graphicx}
\usepackage{amsmath}
\usepackage{amssymb}
\usepackage{multirow}
\usepackage[letterpaper=true,colorlinks,bookmarks=false]{hyperref}
\usepackage{subfigure}
\usepackage[lined,boxed, ruled]{algorithm2e}

\begin{document}

\title{Angle-Closure Detection in Anterior Segment OCT based on Multi-Level Deep Network}

\author{Huazhu~Fu*, Yanwu~Xu*, Stephen~Lin, Damon~Wing~Kee~Wong, Mani~Baskaran, Meenakshi~Mahesh, Tin~Aung, and Jiang~Liu
    \thanks{*Corresponding authors: H.~Fu and Y.~Xu}
    \thanks{H.~Fu is with Inception Institute of Artificial Intelligence, Abu Dhabi, United Arab Emirates, and also with the Institute for Infocomm Research, Agency for Science, Technology and Research, Singapore 138632 (e-mail: huazhufu@gmail.com).}
    \thanks{Y.~Xu is with the Department of Artificial Intelligence Innovation Business, Baidu Inc., Beijing 100193, China (e-mail: ywxu@ieee.org).}
    \thanks{S.~Lin is with Microsoft Research, Beijing 100080, China.}
    \thanks{D.~Wong is with Singapore Eye Research Institute, Singapore 168751, and also with Nanyang Technological University, Singapore 138632.}
    \thanks{M.~Baskaran and M.~Mahesh are with the Singapore Eye Research Institute, Singapore 168751.}
    \thanks{T.~Aung is with Singapore Eye Research Institute, Singapore 168751, and also with the Yong Loo Lin School of Medicine, National University of Singapore, Singapore 119077.}
    \thanks{J.~Liu is with the Cixi Institute of Biomedical Engineering, Chinese Academy of Sciences, Zhejiang 315201, China, and also with the Department of Computer Science and Engineering, Southern University of Science and Technology, Guangdong 518055, China.}}

\markboth{FOR REVIEW}%
{Fu \MakeLowercase{\textit{et al.}}: Regular Paper}

\maketitle

\begin{abstract}

Irreversible visual impairment is often caused by primary angle-closure glaucoma, which could be detected via Anterior Segment Optical Coherence Tomography (AS-OCT). In this paper, an automated system based on deep learning is presented for angle-closure detection in AS-OCT images. Our system learns a discriminative representation from training data that captures subtle visual cues not modeled by handcrafted features. A Multi-Level Deep Network (MLDN) is proposed to formulate this learning, which utilizes three particular AS-OCT regions based on clinical priors: the global anterior segment structure, local iris region, and anterior chamber angle (ACA) patch. In our method, a sliding window based detector is designed to localize the ACA region, which addresses ACA detection as a regression task. Then, three parallel sub-networks are applied to extract AS-OCT representations for the global image and at clinically-relevant local regions. Finally, the extracted deep features of these sub-networks are concatenated into one fully connected layer to predict the angle-closure detection result. In the experiments, our system is shown to surpass previous detection methods and other deep learning systems on two clinical AS-OCT datasets.

\end{abstract}


\IEEEpeerreviewmaketitle

\section{Introduction}

\begin{figure}[!t]
    \begin{center}
        \includegraphics[width=1\linewidth]{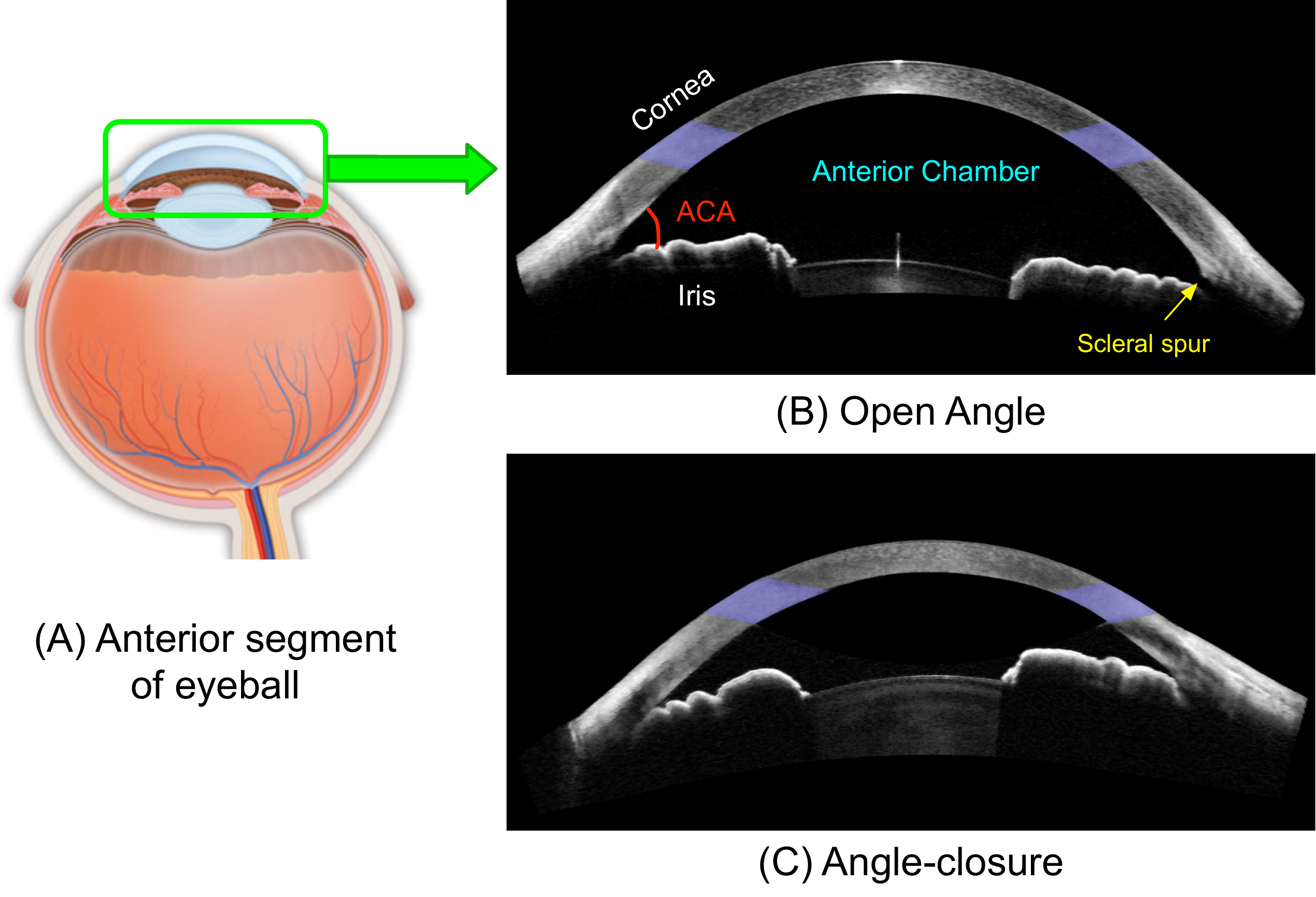}
        \caption{(A)  AS-OCT image captures the anterior segment region of the eye, including cornea, iris, ciliary body and lens. (B) Open angle. (C) Angle-closure.}
        \label{img-cover}
    \end{center}
\end{figure}

The foremost cause of irreversible blindness is glaucoma, with primary angle-closure glaucoma (PACG) being a major cause of blindness in Asia~\cite{Quigley01032006,Tham2014}. Since vision loss from PACG cannot be undone, it is essential to screen people at early asymptomatic stages of the disease to prevent vision loss. In general, gonioscopy is a convenient tool for simple opportunistic case detection in clinics, but imaging is envisaged to reduce this burden of screening for angle closure. Recently, anterior segment optical coherence tomography (AS-OCT) technology has enabled high-resolution imaging of anterior segment structures~\cite{Sharma2014,Ang2018}. In contrast to fundus imaging, AS-OCT imaging provides clear cross-sections of the entire anterior chamber for observation. In a specific type of angle-closure, the anterior chamber angle (ACA) between the iris and cornea is narrow and leads to blockage of drainage channels as in Fig.~\ref{img-cover} (C). This blockage can result in rising eye pressure and optic nerve damage~\cite{Wong2009}.  Previous studies have suggested that several anterior chamber measurements such as the anterior chamber angle, width and lens vault are associated with angle closure~\cite{Nongpiur2013,Nongpiur2017}.  When clinicians use AS-OCT to image the angles for diagnosing angle closure in their patients, the AS-OCT images require a degree of interpretation.  Moreover, AS-OCT imaging has limitations when used in clinical management, because of the lack of automated methods to interpret AS-OCT images for the presence of angle closure.

Several automated assessment methods for wide-scale angle-closure detection using AS-OCT have been studied~\cite{Tian2011,Xu2012,Xu2013,Fu_ASOCT_2017,Fu2018miccai}. A semi-automated system is provided in~\cite{Console2008} to produce various anterior segment clinical measurements, but it needs the user to identify the positions of the Scleral Spurs, making it unsuitable for automated analysis of larger-scale datasets. Among fully automated systems, Tian~\textit{et al.} provide a clinical parameter calculation method for High-Definition OCT (HD-OCT) based on detecting the Schwalbe's line~\cite{Tian2011}. Xu~\textit{et al.} employ visual features directly obtained from the AS-OCT images to classify the glaucoma subtype with a support vector machine (SVM)~\cite{Xu2012,Xu2013}. In~\cite{Fu2016EMBC,Fu_ASOCT_2017}, a label transfer method is proposed to combine segmentation, measurement and detection of AS-OCT structures. However, these methods mainly depend on segmentation to extract handcrafted representations, and then conduct angle-closure detection.
Recently, deep learning techniques have been shown to surpass the performance of handcrafted features in many areas~\cite{SIG-039,LeCun2015,Schmidhuber201585}. For example, Convolutional Neural Networks (CNNs) have led to improved performance in image classification~\cite{Krizhevsky2012}, object detection~\cite{Liu2018deep} and image segmentation~\cite{Long2017_FCN}. For retinal fundus images, there have also been several recent works achieving significant performance gains for diabetic retinopathy~\cite{Gulshan2016,ShuWeiTing2017,Zhao2017vessel}, age-related macular degeneration~\cite{Grassmann2018,AMD2018} and glaucoma~\cite{Fu2018MNet,Fu2018DeNet}. These works inspire our examination of deep learning for angle-closure detection in AS-OCT images, and how clinical priors can be utilized to guide development of a suitable deep learning architecture.

In this paper, a deep learning system is proposed for angle-closure detection that makes use of two clinical observations: 1) the main clinical parameters for angle-closure are measured from different regions of the cornea, iris and anterior chamber, and are evident at different scales~\cite{Wu2011,Nongpiur2013}; 2) the intensities of AS-OCT images could vary among different AS-OCT imaging devices, which could degrade feature extraction and classification. Based on these clinical priors, we introduce a Multi-Level Deep Network (MLDN). In our method, multiple parallel sub-networks jointly learn multi-level representations from the multiple regions/levels known to be informative for angle-closure detection in an AS-OCT image. Furthermore, we also propose an intensity-based data augmentation method to elevate robustness to different AS-OCT imaging modalities. Our contributions can be summarized as follows:
\begin{enumerate}
    \item We propose an automated angle-closure detection system by introducing a deep learning algorithm for the purpose of obtaining predictive representations in AS-OCT images.
    \item A multi-level deep architecture is developed to account for clinical priors at different image areas and levels, in which multi-level features are extracted on different image levels through separate and parallel deep sub-networks.
    \item An efficient and simple ACA detection method is utilized based on sliding-window regression, which formulates the ACA localization task as a regression problem.
    \item Intensity-based data augmentation is presented to effectively handle intensity variations among the different AS-OCT imaging modalities.
    \item Two clinical AS-OCT datasets are collected for evaluation, including the Visante AS-OCT and Cirrus HD-OCT modalities. Experiments show that the proposed MLDN outperforms the previous assessment methods and other deep learning architectures.
\end{enumerate}

The rest of the paper is organized as follows: Sec.~\ref{sec-related} reviews some related works. Sec.~\ref{sec-method} introduces the proposed MLDN method and its main components. Sec.~\ref{sec_exp} presents an evaluation of our method on two AS-OCT datasets, and Sec.~\ref{sec_conclusion} concludes the paper.

\section{Related Work}
\label{sec-related}

\begin{figure*}[!t]
    \begin{center}
        \includegraphics[width=1\linewidth]{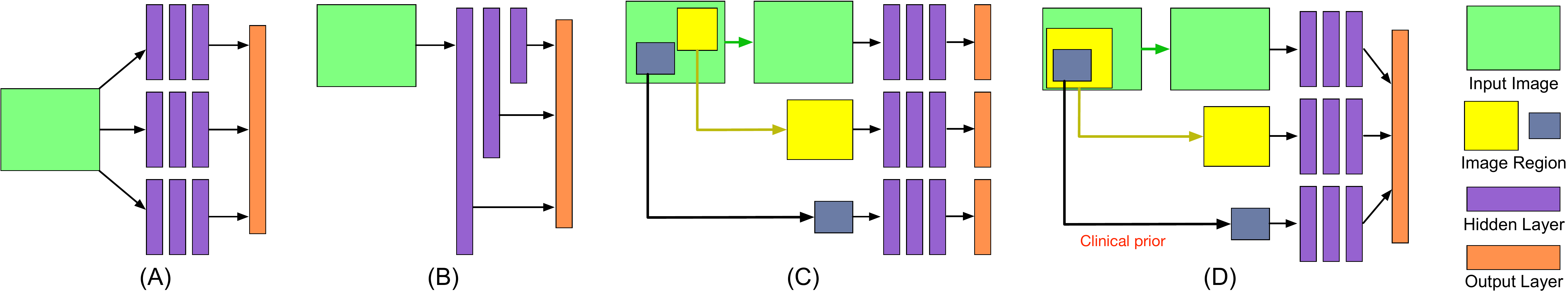}
        \caption{Different deep learning architectural configurations: (A) multi-column architecture; (B) multi-scale architecture; (C) multi-region architecture; (D) our proposed multi-level architecture, where regions are extracted based on clinical guidance. The green block denotes the global image, while the yellow and blue blocks are proposed regions of attention. The purple and orange blocks denote the hidden layers for deep learning (e.g., convolutional layer) and the final output layer (e.g., fully connected layer), respectively.}
        \label{img-deepmodel}
    \end{center}
\end{figure*}

\subsection{Angle-closure detection in AS-OCT image}

The angle-closure detection systems in AS-OCT images can be categorized into two classes. The first is clinical quantitative parameter based, which calculates clinically-relevant measurements based on segmentation results for classification into glaucoma types~\cite{Tian2011,Williams2013,Niwas2015,Fu_ASOCT_2017}. The clinical quantitative parameters are predefined based on anatomical structures~\cite{Leung2008,Wu2011,Nongpiur2013,Tan2012LV} (e.g., anterior chamber width, lens-vault, and trabecular-iris angle), which have explicit clinical meaning used for reference in making a diagnosis. Although such parameters are helpful to clinicians, their measurement depends heavily on the performance of AS-OCT segmentation. A different kind of screening method directly extracts visual features from the AS-OCT images~\cite{Xu2012,Xu2013,NiNi2014}. Visual representations could mine a wider set of image information, and can obtain better screening performance than from clinical parameters. For example, the works in~\cite{Xu2012,Xu2013} employ the Histogram of Oriented Gradients (HOG)~\cite{HOG2005} and histogram equalized pixel (HEP)~\cite{Gangeh2010} with linear SVM to represent the ACA region. An Anterior Chamber Shape feature is proposed in~\cite{NiNi2014} to describes iris shape information. However, hand-crafted visual features such as these may not provide sufficiently discriminative representations. In our work, we present a method based on visual features, but learn a rich hierarchical representation for angle-closure detection in AS-OCT through deep learning.

\subsection{Deep learning}

Based on advances in discriminative representations and large scale data analysis, deep learning techniques have been demonstrated to achieve significant performance gains in many areas, e.g., image classification~\cite{Krizhevsky2012,ResNet2016}, object detection~\cite{RCNN_2014,Redmon_2016_CVPR,Zhu2018YoTube},  image segmentation~\cite{Zheng2015,Long2017_FCN,Fu2016}, and person re-identification~\cite{Zheng2018SIFT,Lei2018}. For learning a rich hierarchical representation, the concept of multi-scale and multi-region networks have been proposed. For example, the multi-column network~\cite{MCNN2012,Zhang2016} has different receptive field sizes for each sub-network, and combines the output maps, as shown in Fig.~\ref{img-deepmodel} (A). The skip-layer network~\cite{Eigen2014,Yang2016} combines features from different pooling layers to build a representation, as shown in Fig.~\ref{img-deepmodel} (B).  Another deep architecture is the multi-region network~\cite{RCNN_2014,Li2015}, which distinguishes object regions/proposals by using individual networks, as shown in Fig.~\ref{img-deepmodel} (C). The target regions are selected by using general object detectors~\cite{Uijlings2013,Arbelaez2011}. Our multi-level network extracts image representations at different levels, but generates a more concise feature focusing both on global and local clinically-relevant regions.  The clinical guidance is introduced to extract appropriate regions for angle-closure detection, as shown in Fig.~\ref{img-deepmodel} (D). Avoiding redundant information in this manner facilitates deep network training.

\section{Proposed Method}
\label{sec-method}

\begin{figure}[!t]
    \begin{center}
        \includegraphics[width=1\linewidth]{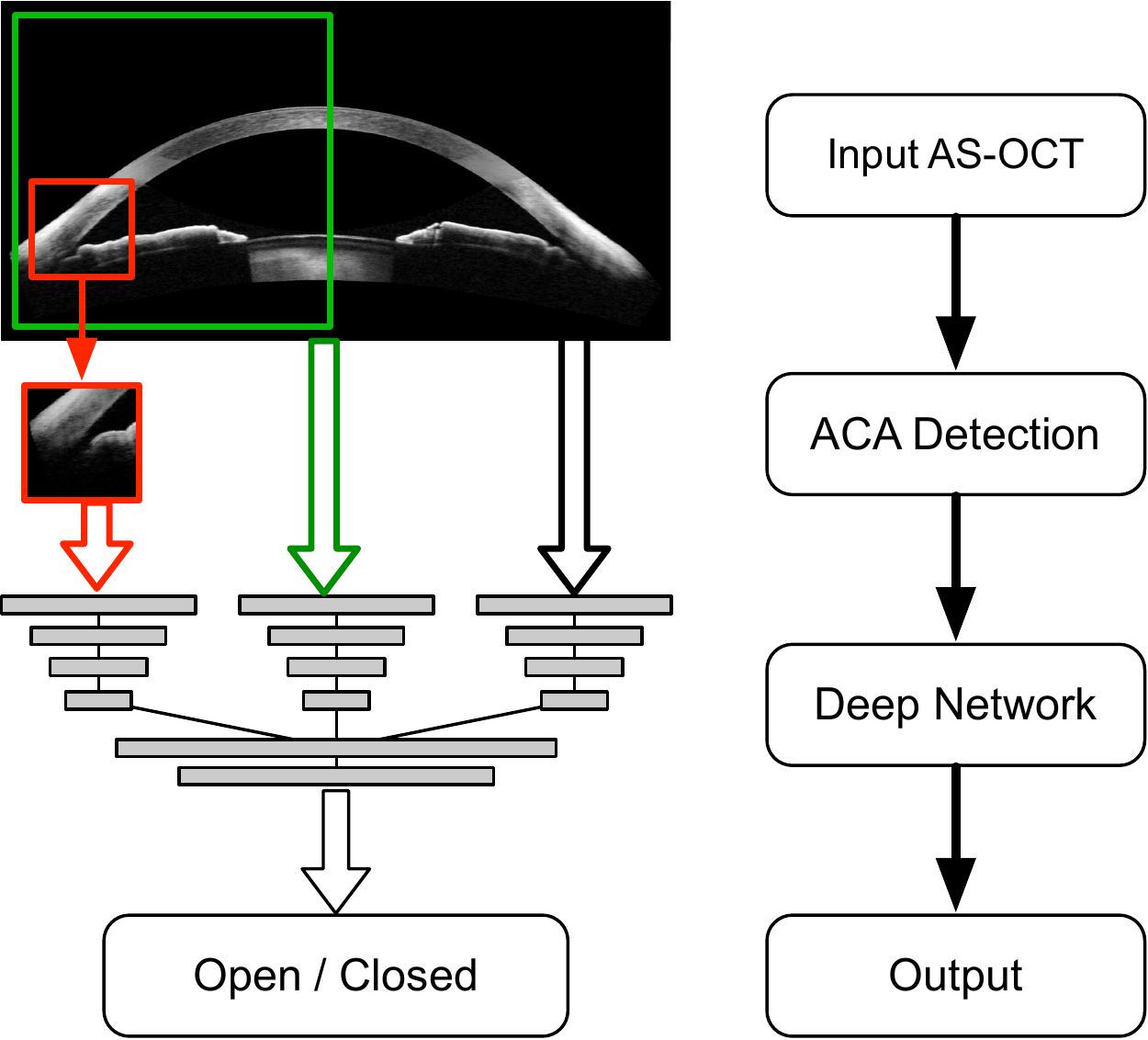}
        \caption{System flowchart.}
        \label{img-flow}
    \end{center}
\end{figure}

In this paper, a Multi-Level Deep Network (MLDN) for automated angle-closure detection in AS-OCT images is proposed. As shown in Fig.~\ref{img-flow}, in our system, the ACA region is detected by using sliding windows, and then the proposed MLDN combines global and local level representations for angle-closure detection in AS-OCT images. Our MLDN consists of three parallel sub-networks to extract hierarchical representations from different clinically-relevant regions in an AS-OCT image, including the global anterior segment structure, local iris region, and ACA patch. Finally, the output maps of the three sub-networks are concatenated as input into one fully connected layer to predict the angle-closure detection result.

\subsection{ACA region detection}

\begin{figure*}[!t]
    \begin{center}
        \includegraphics[width=1\linewidth]{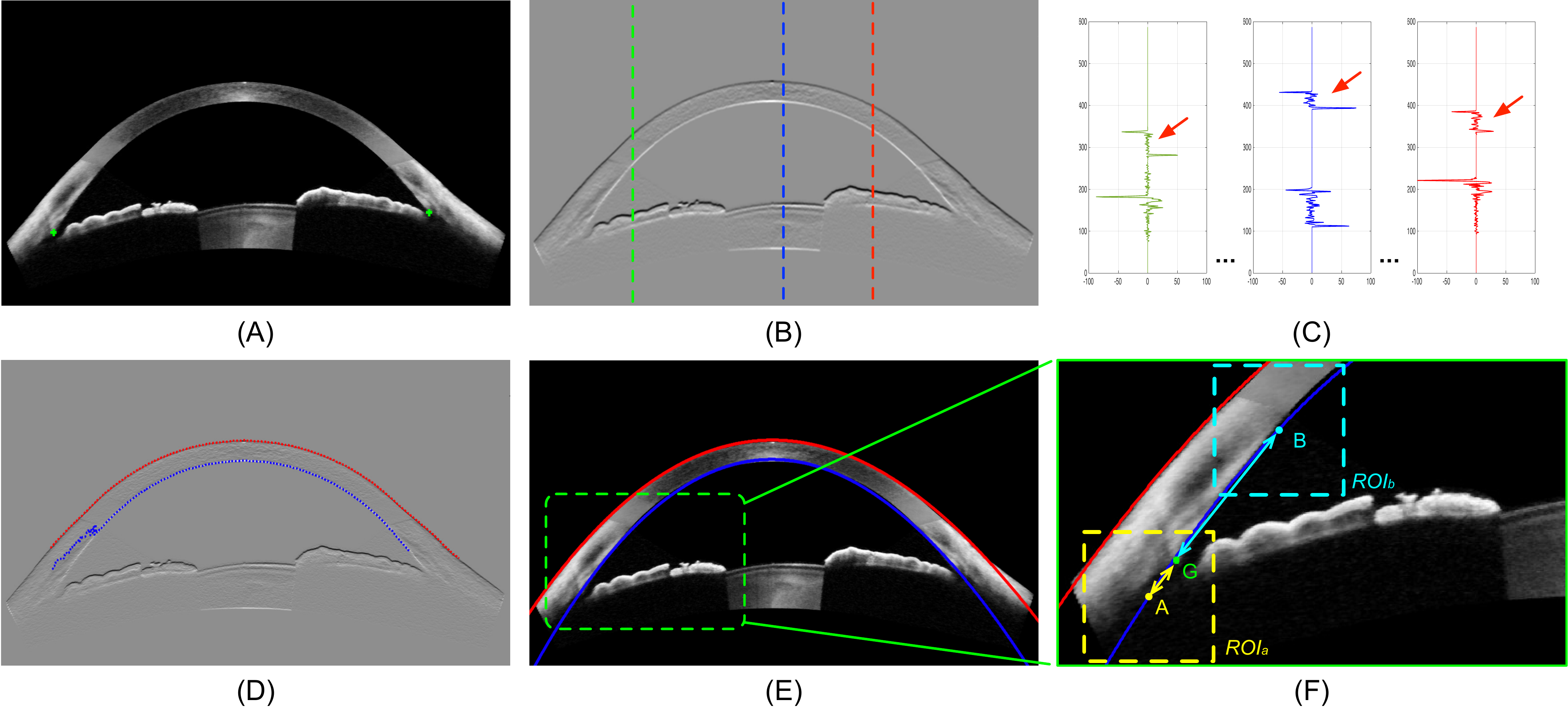}
        \caption{(A) Original AS-OCT with the Scleral Spur points highlighted in green. (B) The vertical gradient map of an AS-OCT image. (C) The gradient value responses for the marked columns in the gradient map (B). The red arrow indicates the local impulse response corresponding to the corneal boundary. (D) The estimated initial points of the corneal boundary. (E) AS-OCT with the estimated corneal boundaries. The red and blue dotted curves denote the estimated corneal boundaries. (F) A zoom-in region of the AS-OCT image. Two sliding ROIs are marked by yellow and cyan rectangles. The distance between the ROI center (e.g., $A, B$) to the Scleral Spur point $G$ is calculated and used for regression.}
        \label{img-aca_region}
    \end{center}
\end{figure*}

The Anterior Chamber Angle (ACA) lies at the junction of the cornea and the iris, as shown in Fig.~\ref{img-cover} (B), and is a major bio-marker in AS-OCT images, providing key risk indicators for angle closure. Several methods have been proposed to detect the ACA region. For example, the work in~\cite{Xu2012} proposed a geometric method based on edge detection and shape properties to detect the ACA region. But this method only works well for certain AS-OCT imagery, and the quality of image will greatly affect performance. Tian \textit{et al.} provided an algorithm to localize the ACA for High-Definition OCT (HD-OCT) images by using the Schwalbe's line~\cite{Tian2011}. However, it is limited to HD-OCT images and not valid for low resolution modalities like AS-OCT. An automated AS-OCT structure segmentation based on label transfer was presented in~\cite{Fu_ASOCT_2017}, which transfers manually marked labels to the target image for guiding segmentation. In our work, we only require localization of the ACA region without pixel-level segmentation. We instead employ a simple but effective method based on a sliding window regression framework. In our method, we first segment the corneal boundary, and then a series of ROIs are extracted along the corneal boundary as regression candidates. The aim is to detect the Scleral Spur (SS) as a landmark, and crop a small region of interest (ROI) centered on the detected SS as the ACA patch.

Given the AS-OCT image, we first segment the corneal boundary, which is the top curve of the anterior segment region, as shown in Fig.~\ref{img-aca_region} (A). After reducing speckle noise with a 2-D Gaussian filter, we calculate the vertical gradient map of the AS-OCT image, as shown in Fig.~\ref{img-aca_region} (B). The corneal upper and bottom boundaries appear at the top negative and positive impulse responses in each column of the gradient map, respectively, as shown in Fig.~\ref{img-aca_region} (C). We estimate the initial points by finding the top two edge points for each column in the AS-OCT, as shown in Fig.~\ref{img-aca_region} (D), and then two fourth-order polynomials~\cite{Fu_ASOCT_2017} are fitted to estimate the corneal boundary as shown in Fig.~\ref{img-aca_region} (E).

With the detected corneal boundary, we sample a series of ROIs (with a size of $120 \times 120$ in our paper)  as candidates along the corneal bottom boundary to localize the SS point, which is formulated as a regression task, as shown in Fig.~\ref{img-aca_region} (F). For regression, the distance between the ROI center to the SS point is used:
\begin{equation}
d = \min\left\{1, \dfrac{2 |v_r - v_s|}{W_r}\right\} ,
\end{equation}
where the $v_r$ and $v_s$ denote the horizontal coordinates of the ROI center and SS point, respectively. $W_r$ is the width of the ROI. The $\min \{1, *\}$ ensures that the distance lies in the range of $[0 ,1]$ when the SS point is outside of the ROI. In our method, we employ L2-regularized support vector regression (SVR)~\cite{liblinear}:
\begin{equation}
\min_{\textbf{w}_r} \dfrac{1}{2} \textbf{w}^T_r \textbf{w}_r + C \sum_{i=1}^{l}  (\max(0, |\textbf{w}^T_r \textbf{x}_i - d_i| - \epsilon)^2) ,
\end{equation}
where $C > 0$ is a regularization parameter, $\textbf{w}_r$ is the regression parameter vector, and $l$ is the training number. The parameter $\epsilon$ is employed so that the loss is zero if $|\textbf{w}^T_r \textbf{x}_i - d_i|\leq \epsilon$. $(\textbf{x}_i, d_i)$ are the ROI-label pairs with the feature vector $\textbf{x}_i$ and corresponding distance $d_i$ of ROI $i$. For ACA detection, we repeat the steps, and output the ROI with the smallest regression value as the ACA patch. In this paper, we implement the SVR model based on the liblinear toolbox~\cite{liblinear}, and the HOG~\cite{HOG2005} feature is utilized to represent the ROI.

\subsection{Network architecture}

\begin{figure*}[!t]
    \begin{center}
        \includegraphics[width=1\linewidth]{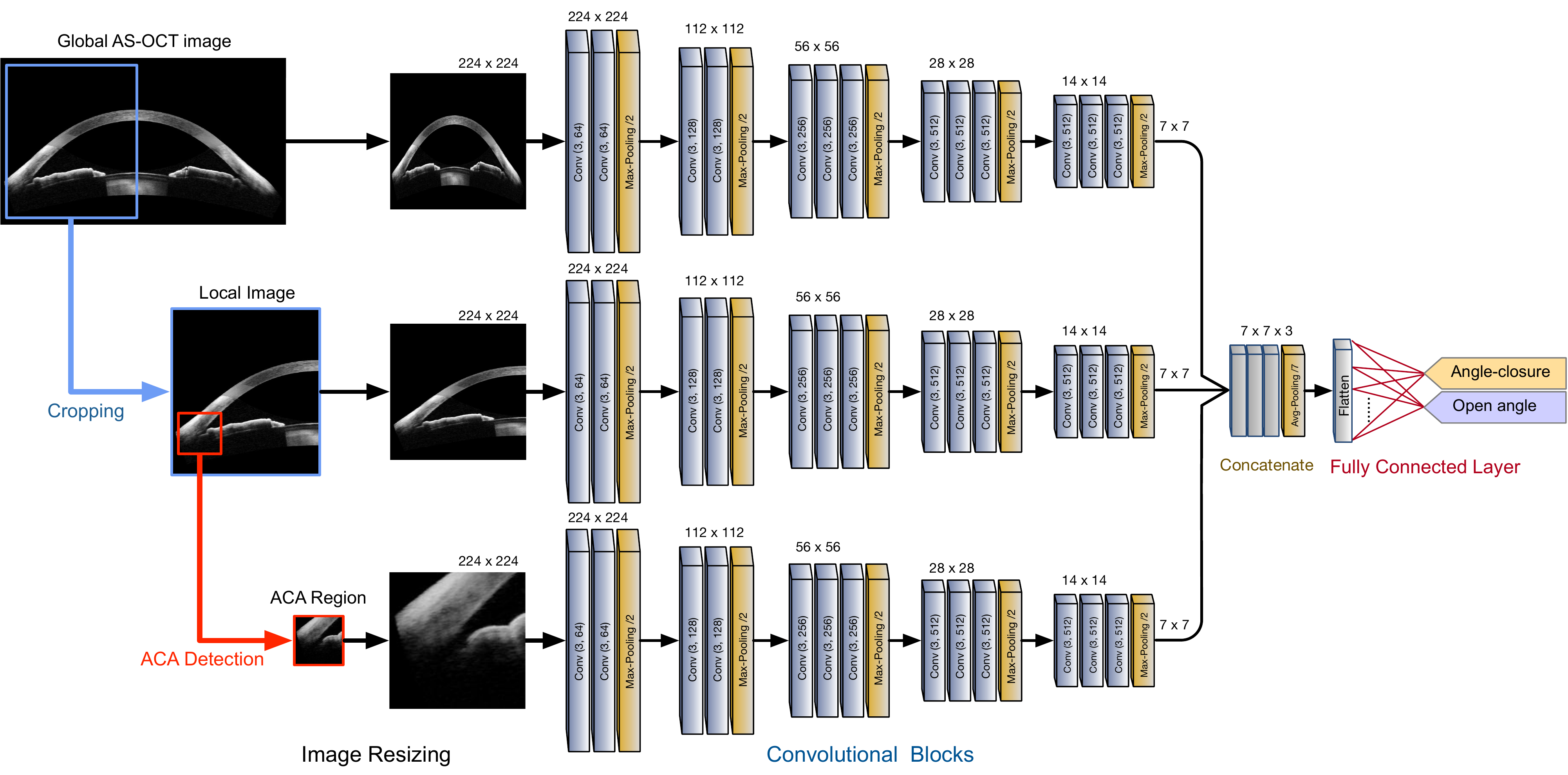}
        \caption{Architecture of our multi-level deep network. Each convolutional block contains a convolutional layer, batch normalization, and ReLU activation. The merge layer (orange block) combines the feature maps of the three sub-networks, while one fully connected layer with softmax is used to predict the detection result. The parameters of the convolutional block are denoted as "Conv (channel number, kernel size)"}.
        \label{img-frame}
    \end{center}
\end{figure*}

Our proposed Multi-Level Deep Network (MLDN) consists of three parallel sub-networks to generate representations for multiple clinically-relevant regions in an AS-OCT image, as shown in Fig.~\ref{img-frame}, including the global anterior segment structure, local iris region, and ACA patch.
Each sub-network contains a sequence of convolutional blocks composed of a convolutional layer, batch normalization~\cite{IoffeBN}, and ReLU activation~\cite{Krizhevsky2012}. In them, the convolutional layers learn and generate local feature maps based on local receptive fields over its input, while the batch normalization acts as a regularization that helps to avoid overfitting~\cite{IoffeBN}. The ReLU is employed as the activation function following the convolutional layer.

The first sub-network of our MLDN works on the global AS-OCT image to represent the whole anterior segment structure. The anterior segment structure provides multiple clinical measurements~\cite{Wu2011}, such as lens vault and anterior chamber structure. The second sub-network focuses on the local region containing the iris structure, as iris-related parameters (e.g., angle recess area, anterior chamber depth, iris shape and curvature) are widely used in clinical diagnosis of angle-closure~\cite{NiNi2014}. This local region is obtained by cropping half of the AS-OCT image containing only one ACA region.
The last sub-network processes a small patch centered on the ACA area. In clinical diagnosis, the ACA area is the most important because several major risk factors for angle-closure detection are calculated from it~\cite{Nongpiur2013} (e.g., trabecular-iris angle and angle opening distance).
For each sub-network, we resize the input images to $224 \times 224$ to implement other pre-trained deep models conveniently~\cite{Krizhevsky2012,Simonyan14c}.

The feature maps of the three sub-networks are combined and fed into one fully connected layer with softmax regression to predict the final detection result. As a binary classification task, the binary cross-entropy is utilized as the loss function for angle-closure detection:
\begin{equation}
L = - \dfrac{1}{N} \sum_{i=1}^{N} [y_i \log (p_i) + (1-y_i) \log (1- p_i) ],
\end{equation}
where $p_i$ is estimated probability for the $i$-th image, $y_i \in \{0,1\}$ is the angle-closure label, and $N$ denotes the number of training images.

\subsection{Intensity-based data augmentation}

The AS-OCT images are captured along the perpendicular direction of the eye, and anterior chamber structures generally appear at consistent positions in AS-OCT images. As a result, general image augmentation operations by rotation and scaling do not facilitate training for AS-OCT images. Moreover, the different AS-OCT modalities may lead to varied image intensities, which can degrade detection performance. Based on this, an intensity-based data augmentation is introduced on the training data, specifically by re-scaling image intensities by a factor of $k_I$ ($k_I=[0.5, 1, 1.5]$ in this paper). We note that errors may exist in the localization of the ACA patch. To reduce the effect of such errors, we also augment the data by shifting the ACA center to generate different patches for input into the ACA sub-network.

\section{Experiments}
\label{sec_exp}

\subsection{Clinical AS-OCT dataset}

For experimentation, we collected two clinical AS-OCT datasets for different devices, namely a Carl Zeiss Visante AS-OCT and a Cirrus HD-OCT.

\begin{figure}[!t]
    \begin{center}
        \includegraphics[width=1\linewidth]{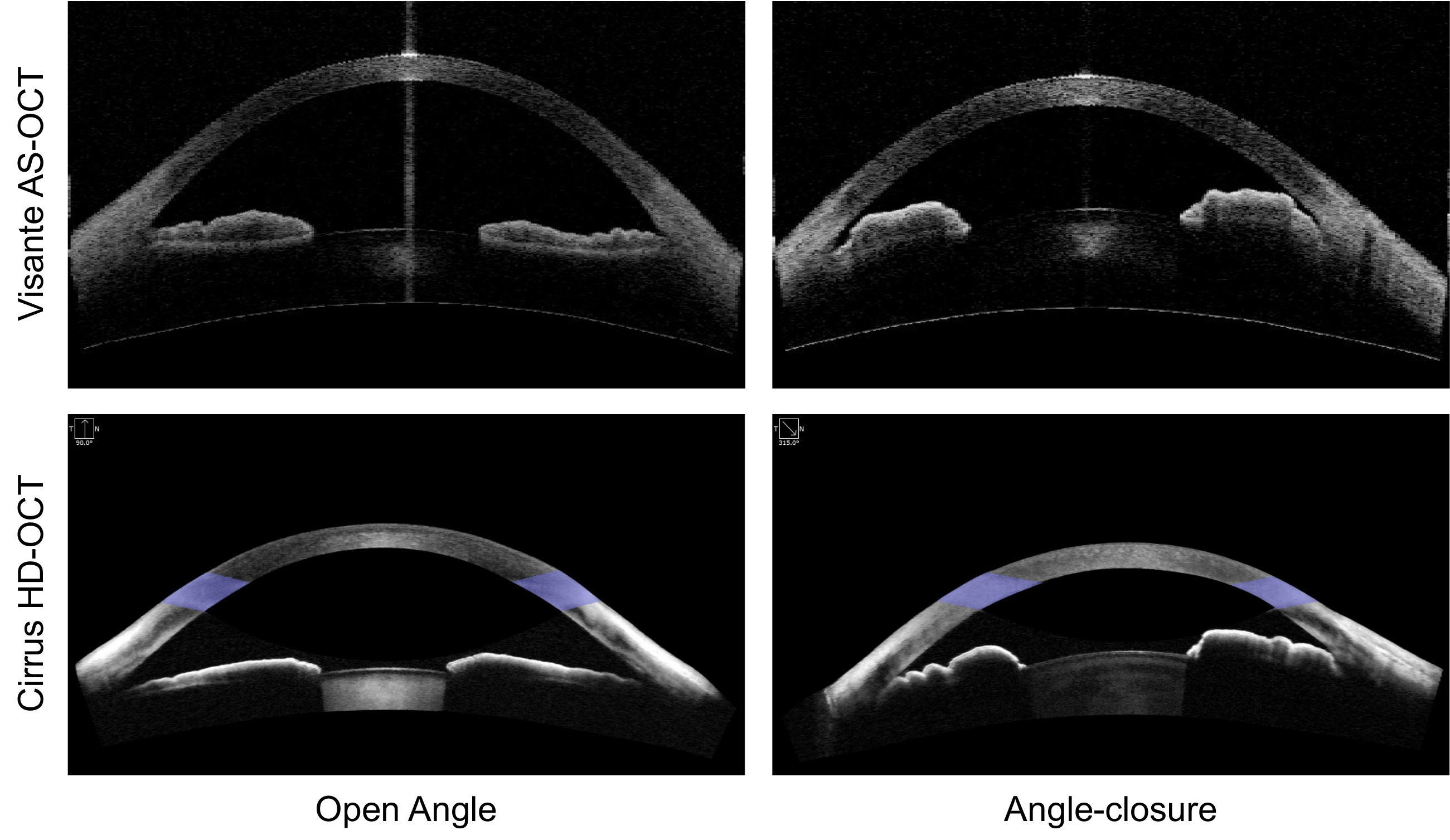}
        \caption{Example images from the two AS-OCT datasets.}
        \label{img-exp_data}
    \end{center}
\end{figure}

\textbf{Visante AS-OCT dataset} is captured using a Visante AS-OCT machine (Model 1000, Carl-Zeiss Meditec). The dataset is composed of 4135 AS-OCT images from 2113 subjects. As two ACA regions exist in one AS-OCT image, we split each AS-OCT into two ACA images (yielding a total of 8270 ACA images), and flip the right-side ACA images horizontally to align with the left-side images. The ground truth for each ACA image is labeled as open-angle or angle-closure. It is defined as the presence (angle-closure) or absence (open-angle) of iris-trabecular contact by visual inspection of the images by a clinician trained in examining the images, which are later adjudicated by a senior ophthalmologist trained and experienced in glaucoma diagnosis. Overall, the whole dataset contains 7375 open-angle and 895 angle-closure ACA images.

\textbf{Cirrus HD-OCT dataset} is captured using a Cirrus HD-OCT machine (Model HD 5000, Version 8.0, Carl-Zeiss Meditec, Germany). It contains 701 AS-OCT images in total from 202 subjects. Same as with Visante AS-OCT, two ACA images are extracted from each AS-OCT image, and finally we obtain 1102 open-angle and 300 angle-closure ACA images. Different from the Visante AS-OCT imaging, post-processing is done during Cirrus machine imaging, where the corneal boundary is made clearer, as shown in Fig.~\ref{img-exp_data}. However, the posterior border of the iris and the iris trabecular contact points are not obvious in all images.

In the experiment, each dataset is divided randomly into training and testing sets at a patient level, which ensures that the ACA images from one patient fall into the same training or test set.

\subsection{Implementation}

Training of the proposed MLDN is composed of two phases. First, the pre-trained model trained from ImageNet is loaded in each sub-network as the initialization, and then each sub-network is fine-tuned individually on the training dataset. Second, we combine the three trained sub-networks together into the MLDN and fine-tune the last fully connected layer on the training dataset. To optimize our deep models, stochastic gradient descent (SGD) is conducted on mini-batches with a learning rate of $1e-4$ and a momentum of $0.9$. The whole framework is implemented by using Keras with Tensorflow. Fine-tuning (200 epochs) requires approximately five hours on one NVIDIA K40 GPU.  In testing, the detection result is produced within 500 ms per AS-OCT image including ACA detection and classification.

\subsection{Experimental criterion and baseline}

For detection evaluation, Receiver Operating Characteristic (ROC) curves and area under ROC (AUC) curve by applying different thresholds are reported. Moreover, we also represent the highest point on the ROC curve as the diagnostic threshold to calculate four evaluation criteria, namely Sensitivity ($Sen$), Specificity ($Spe$), Balanced Accuracy ($BAcc$), and F-measure ($F_m$):
\begin{align}
& Sen = \dfrac{TP}{TP+FN}, \; Spe = \dfrac{TN}{TN+FP}, \nonumber \\
& BAcc = \dfrac{1}{2}(Sen + Spe), \;  F_m = \dfrac{2 TP}{2TP + FP +FN}, \nonumber
\end{align}
where $TP$, $TN$, $FP$ and $FN$ denote the number of true positives, true negatives, false positives and false negatives, respectively.

We compare our algorithm with some angle-closure detection methods and deep learning networks:

\textbf{- Clinical parameter based method}~\cite{Fu_ASOCT_2017}, which segments the AS-OCT image by using the label transfer technique and calculates the pre-defined clinical parameters. The screening result is then obtained from the clinical parameters via a linear SVM.

\textbf{- Visual feature based method}~\cite{Xu2012}, which detects the ACA region based on corneal structure and employs HOG features for classifying the angle-closure. The HOG feature is extracted from a $150 \times 150$ patch centered on the ACA region, and a linear SVM is used to predict the detection result.

\textbf{- Single-network Deep Model.} Three state-of-the-art deep learning architectures are used as baselines: VGG-16~\cite{Simonyan14c}, Inception-V3~\cite{Szegedy_2016_CVPR}, and ResNet50~\cite{ResNet2016}. VGG-16 has 13 convolutional layers and 3 fully connected layers~\cite{Simonyan14c}. Inception-V3 utilizes factorized smaller convolutions and aggressive spatial regularization to avoid representational bottlenecks~\cite{Szegedy_2016_CVPR}. ResNet50 employs a residual function to gain accuracy from considerably increased depth~\cite{ResNet2016}. For each deep model, we report three results: one is based on the original global image (Img), the second is local iris region (Loc) as used in our second sub-network, and the last is based on the ACA patch (ACA) used in our third sub-network.

\textbf{- Multi-Scale Deep Model:} We also report results from two types of multi-scale deep learning models. The first is the multi-column deep model used in~\cite{MCNN2012,Zhang2016} with different receptive field sizes for each sub-network, as shown in Fig.~\ref{img-deepmodel} (A). The second is the skip-layer deep model~\cite{Eigen2014,Yang2016}, which combines different pooling layers to build the representation on a primary sub-network, as shown in Fig.~\ref{img-deepmodel} (B).

\subsection{Detection evaluation}

\begin{table*}[!t]
    \renewcommand{\arraystretch}{1.2}
    \centering
    \caption{Performance of different methods on two datasets. The best two results are shown in \textcolor{red}{red} and \textcolor{blue}{blue}, respectively.}
    \begin{tabular}{|l||c|c|c|c||c|c|c|c|}
        \hline
                               &                               \multicolumn{4}{c||}{Visante AS-OCT Dataset}                                &                                \multicolumn{4}{c|}{Cirrus HD-OCT Dataset}                                 \\ \hline
        Methods                &        B-Accuracy        &       Sensitivity        &       Specificity        &        F-measure         &        B-Accuracy        &       Sensitivity        &       Specificity        &        F-measure         \\ \hline
        Quantitative Parameter &          0.8186          &          0.7914          &          0.8459          &          0.5058          &          0.7824          & \textcolor{blue}{0.8559} &          0.7090          &          0.5026          \\
        Visual Feature         &          0.8667          &          0.8776          &          0.8558          &          0.5609          &          0.8299          &         {0.7748}         &          0.8849          &          0.6491          \\ \hline
        VGG-16 Img             &          0.8823          &          0.8957          &         {0.8689}         &          0.5904          &          0.8465          &          0.7928          &          0.9002          &          0.6822          \\
        VGG-16 Loc             &          0.8895          &          0.8685          &         {0.9104}         &          0.6564          &          0.8357          & \textcolor{blue}{0.8559} &         {0.8156}         &          0.6032          \\
        VGG-16 ACA             &         {0.9003}         &         {0.8980}         &          0.9026          &         {0.6545}         & \textcolor{blue}{0.8792} &          0.8378          &         {0.9205}         &         {0.7410}         \\
        Inception-V3 Img       &          0.8771          &          0.8662          &          0.8880          &          0.6102          &          0.7698          &          0.7748          &          0.7648          &          0.5119          \\
        Inception-V3 Loc       &          0.8998          &          0.9070          &          0.8927          &          0.6390          &          0.7751          &          0.7838          &          0.7665          &          0.5179          \\
        Inception-V3 ACA       &          0.9008          &          0.8889          & \textcolor{blue}{0.9128} &          0.6718          &          0.8690          &          0.8108          &          0.9272          &          0.7377          \\
        ResNet50 Img           &          0.8269          &          0.7664          &          0.8874          &          0.5587          &          0.7453          &          0.6937          &          0.7970          &          0.4968          \\
        ResNet50 Loc           &          0.8462          &          0.8209          &          0.8715          &          0.5591          &          0.7276          &          0.6937          &          0.7614          &          0.4681          \\
        ResNet50 ACA           &          0.8781          &          0.8481          &         {0.9081}         &          0.6410          &          0.7847          &          0.8198          &          0.7496          &          0.5200          \\
        Multi-Column Model     & \textcolor{blue}{0.9056} & \textcolor{blue}{0.9274} &          0.8838          &          0.6317          &         {0.8595}         &         {0.7748}         & \textcolor{blue}{0.9442} & \textcolor{blue}{0.7478} \\
        Skip-Layer Model       &         {0.9039}         &         {0.8844}         & \textcolor{red}{0.9235}  & \textcolor{red}{0.6940}  &          0.8602          & \textcolor{blue}{0.8559} &          0.8646          &          0.6643          \\ \hline
        Our MLDN               & \textcolor{red}{0.9180}  & \textcolor{red}{0.9297}  &          0.9062          & \textcolor{blue}{0.6777} & \textcolor{red}{0.9124}  & \textcolor{red}{0.8739}  & \textcolor{red}{0.9509}  & \textcolor{red}{0.8186}  \\ \hline
    \end{tabular}%
    \label{Tab_result}%
\end{table*}%

\begin{figure*}
    \centering
    \subfigure[]
    {\includegraphics[width=.47\linewidth, height= .4\linewidth]{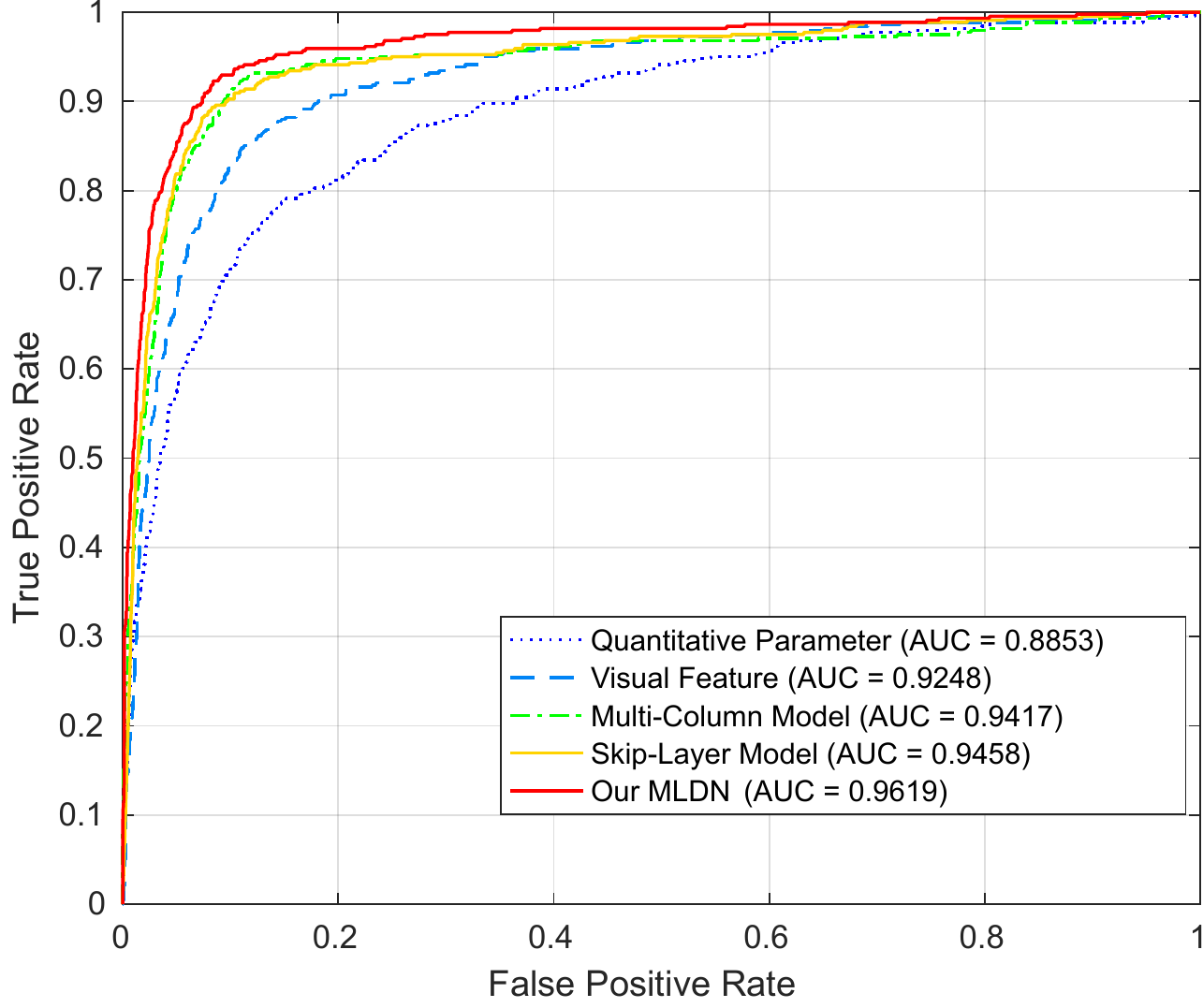}
        \label{exp_Visante_base}} \hfill
    \subfigure[]
    {\includegraphics[width=.47\linewidth, height= .4\linewidth]{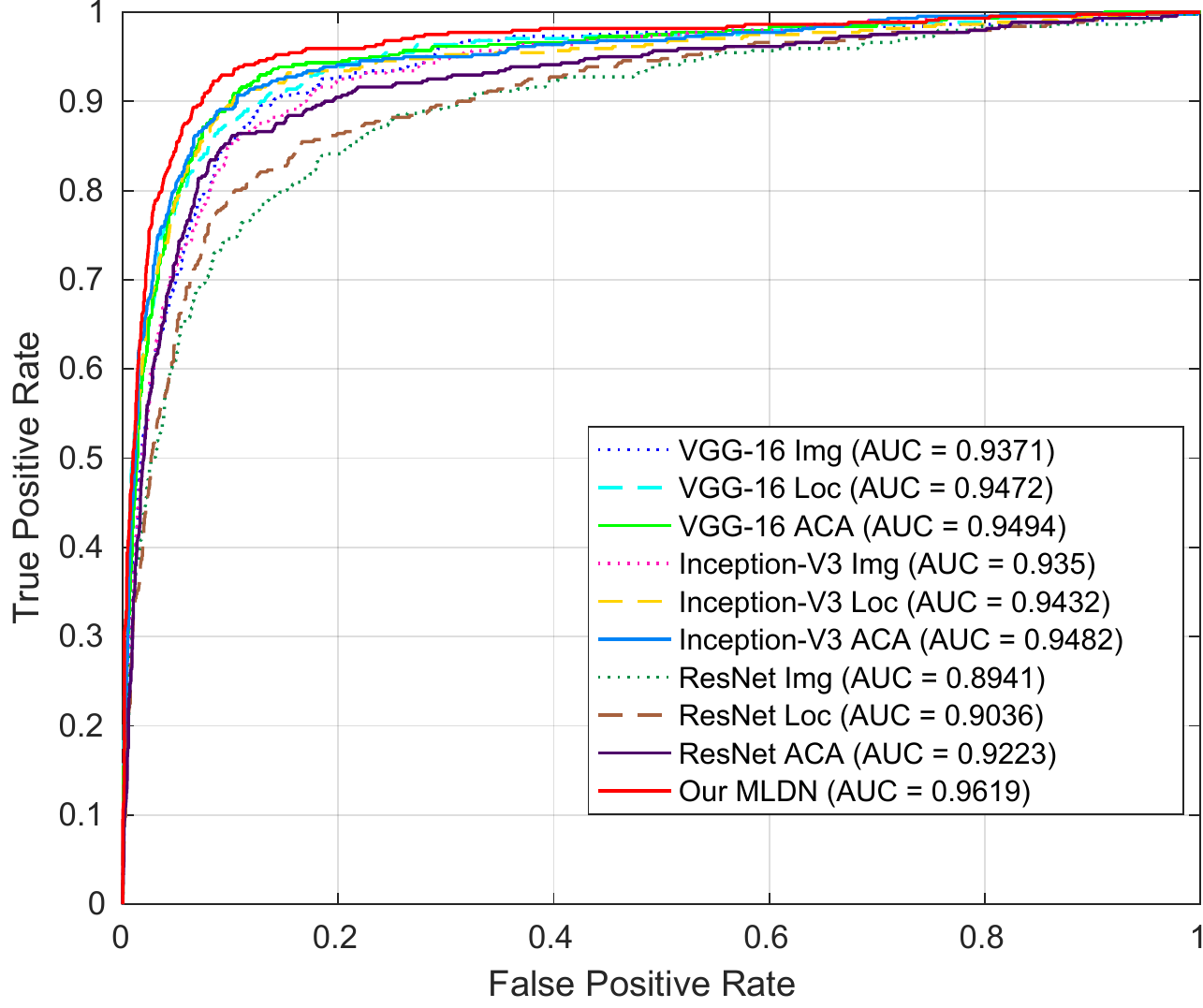}
        \label{exp_Visante_deep}}
    \\
    \subfigure[]
    {\includegraphics[width=.47\linewidth, height= .4\linewidth]{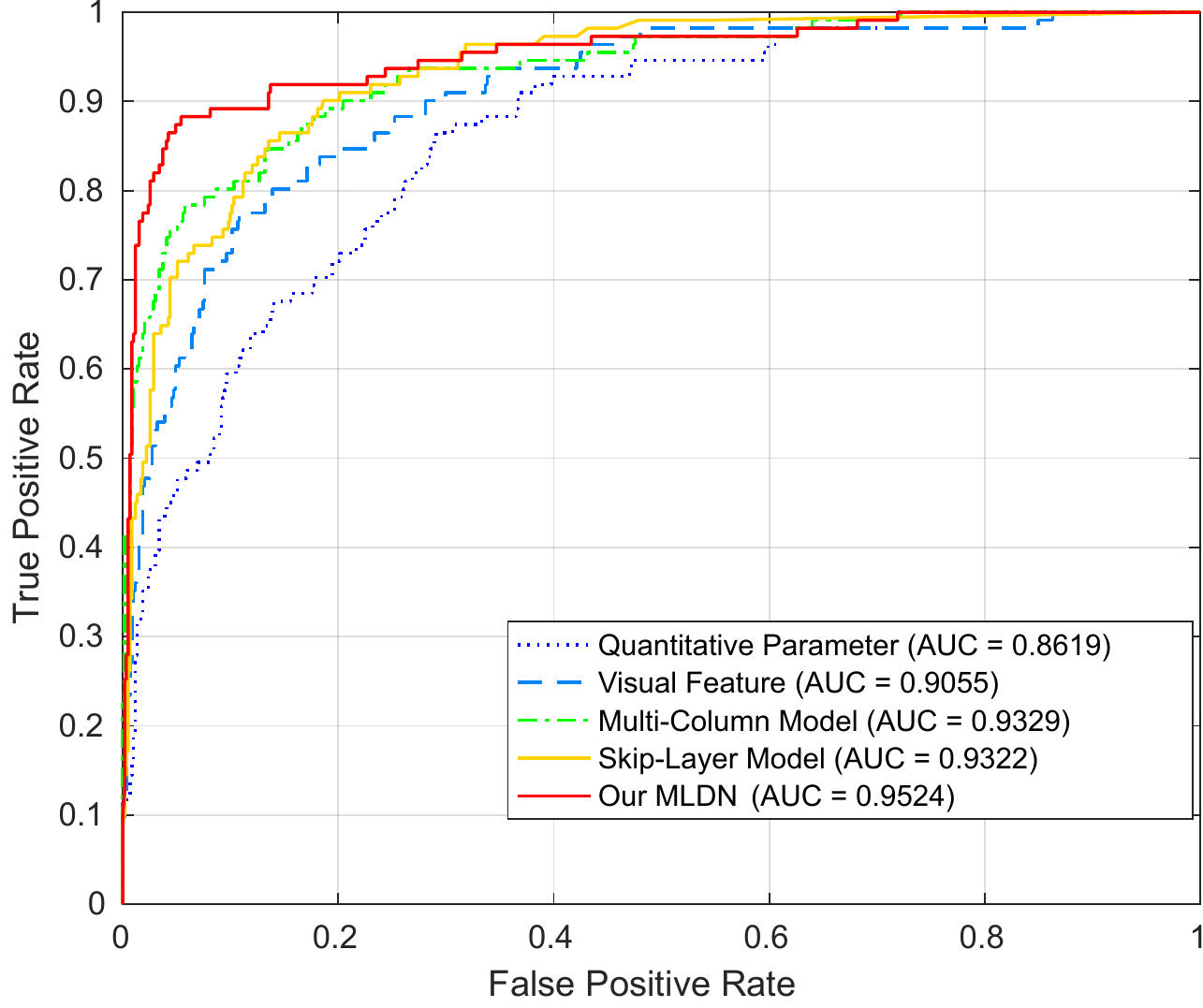}
        \label{exp_Cirrus_base}} \hfill
    \subfigure[]
    {\includegraphics[width=.47\linewidth, height= .4\linewidth]{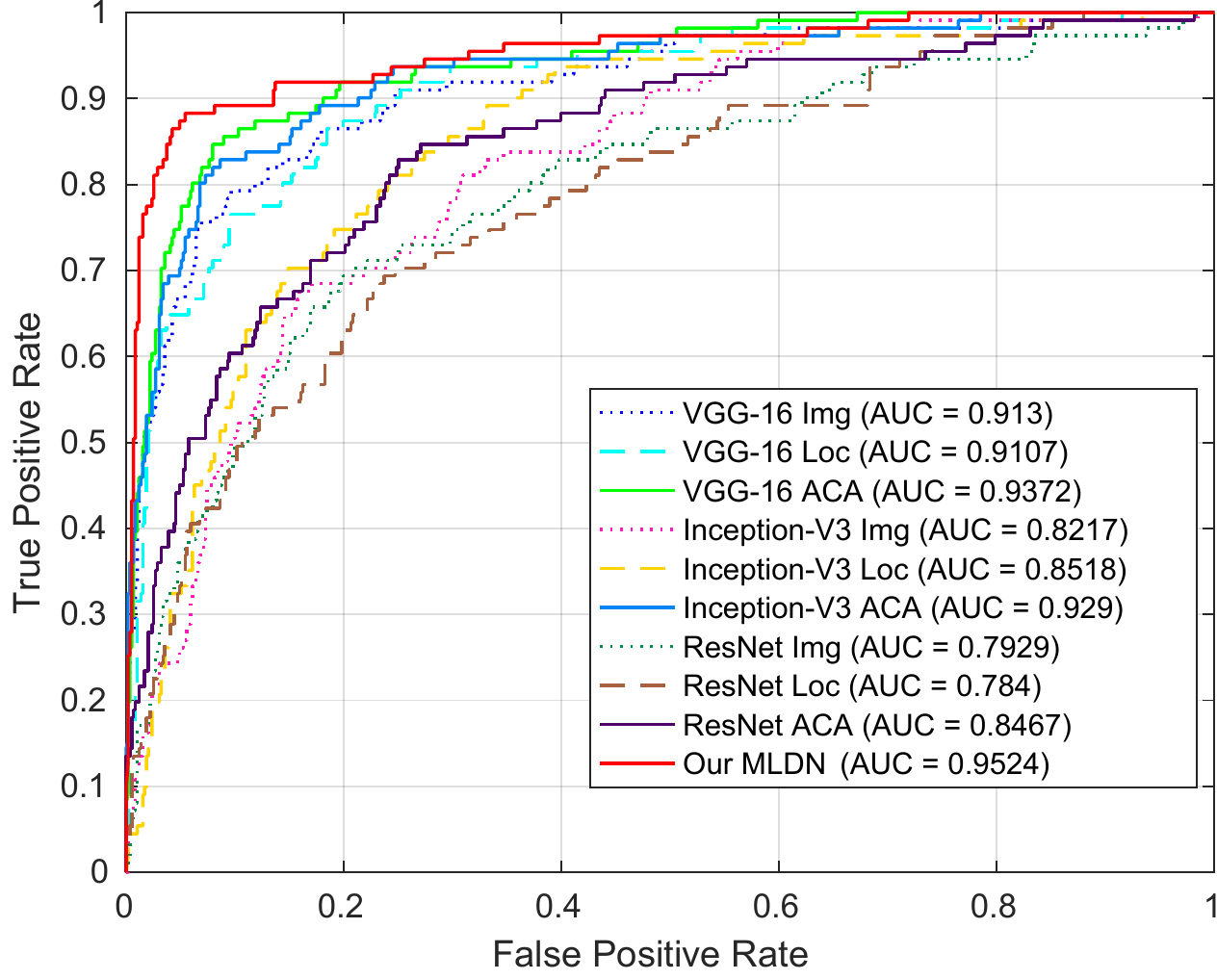}
        \label{exp_Cirrus_deep}} \\
    \caption { Performance of different methods on the (A)-(B) Visante AS-OCT dataset and (C)-(D) Cirrus HD-OCT dataset. }
    \label{exp_curve}
\end{figure*}

We test our MLDN on the Visante and Cirrus datasets. The performances are reported in Table~\ref{Tab_result} and Fig.~\ref{exp_curve}. Most clinical measurements are defined according to anatomical structures, which are known to have specific physical significance in making a diagnosis. However, the clinical parameters do not represent information beyond what clinicians observed, and do account for detailed visual context of the AS-OCT. Thus the performance with clinical parameters does not perform best in automated screening based on AS-OCT. Visual representations could mine more image details, and is found to achieve better scores than the clinical parameter based method as expected. This was also observed in previous work~\cite{Fu_ASOCT_2017}.

The deep learning based methods can learn rich representations more powerful than clinical parameters and handcrafted visual features. In the Visante AS-OCT dataset, the deep models (e.g., VGG-16 and Inception-V3) perform better than visual features at all scales (Img, Loc, and ACA). But in the Cirrus HD-OCT dataset, only the performance of the VGG-16 model on Loc and ACA are higher than for visual features. A possible reason for this is that deeper architectures such as Inception-V3 and ResNet require larger-scale datasets for effective fine-tuning, while in our case only smaller-sized datasets (e.g., Cirrus HD-OCT dataset) are available. For different scales of inputs, the ACA patch focuses on the most highly relevant details, and in all the deep models, the results on the ACA patch are higher than for the global image or local iris region. For multi-scale deep models (Multi-Column and Skip-Layer models), the shallow layers can be easier to optimize, resulting in similar performance on both datasets. Our MLDN outperforms the other networks, showing that the proposed combination of multi-level clinically-relevant regions improves the detection performance, with AUC of 0.9619 ($95\%$ Confidence Interval (CI): 0.9499-0.9711) on the Visante AS-OCT dataset and 0.9524 ($95\%$ CI: 0.9208-0.9718) on the Cirrus HD-OCT dataset, respectively. Note that the AS-OCT images were captured with the standardized position and direction, which leads them to have similar structure locations. Alignment pre-processing is thus not needed. Moreover, the data augmentation of the deep learning method promotes robustness of our method against slight shifts and rotations.

\subsection{Component evaluation}

\begin{table*}[!t]
    \renewcommand{\arraystretch}{1.2}
    \centering
    \caption{Performance of different sub-network combinations on two datasets.}
    \begin{tabular}{|l||c|c|c|c||c|c|c|c|}
        \hline
                          &    \multicolumn{4}{c||}{Visante AS-OCT Dataset}    &     \multicolumn{4}{c|}{Cirrus HD-OCT Dataset}     \\ \hline
        Methods           & B-Accuracy & Sensitivity & Specificity & F-measure & B-Accuracy & Sensitivity & Specificity & F-measure \\ \hline
        MLDN (Img)        &   0.8969   &   0.8617    &   0.9321    &  0.7031   &   0.7786   &   0.7297    &   0.8274    &  0.5510   \\
        MLDN (Loc)        &   0.8958   &   0.8798    &   0.9117    &  0.6650   &   0.8292   &   0.8649    &   0.7936    &  0.5836   \\
        MLDN (ACA)        &   0.8932   &   0.8753    &   0.9112    &  0.6615   &   0.8476   &   0.7748    &   0.9205    &  0.7049   \\
        MLDN (Img + Loc)  &   0.9090   &   0.8934    &   0.9245    &  0.7011   &   0.8822   &   0.8829    &   0.8816    &  0.7025   \\
        MLDN (Img + ACA)  &   0.9090   &   0.8934    &   0.9245    &  0.7011   &   0.8732   &   0.8378    &   0.9086    &  0.7209   \\
        MLDN (Loc + ACA)  &   0.9028   &   0.9025    &   0.9031    &  0.6579   &   0.8941   &   0.8829    &   0.9052    &  0.7396   \\
        MLDN (All) w/o AG &   0.9024   &   0.9116    &   0.8932    &  0.6422   &   0.8690   &   0.8919    &   0.8460    &  0.6578   \\
        MLDN  (All)       &   0.9180   &   0.9297    &   0.9062    &  0.6777   &   0.9124   &   0.8739    &   0.9509    &  0.8186   \\
         \hline
    \end{tabular}%
    \label{Tab_Sub_result}%
\end{table*}%

\begin{figure*}
    \centering
    \subfigure[]
    {\includegraphics[width=.47\linewidth, height= .4\linewidth]{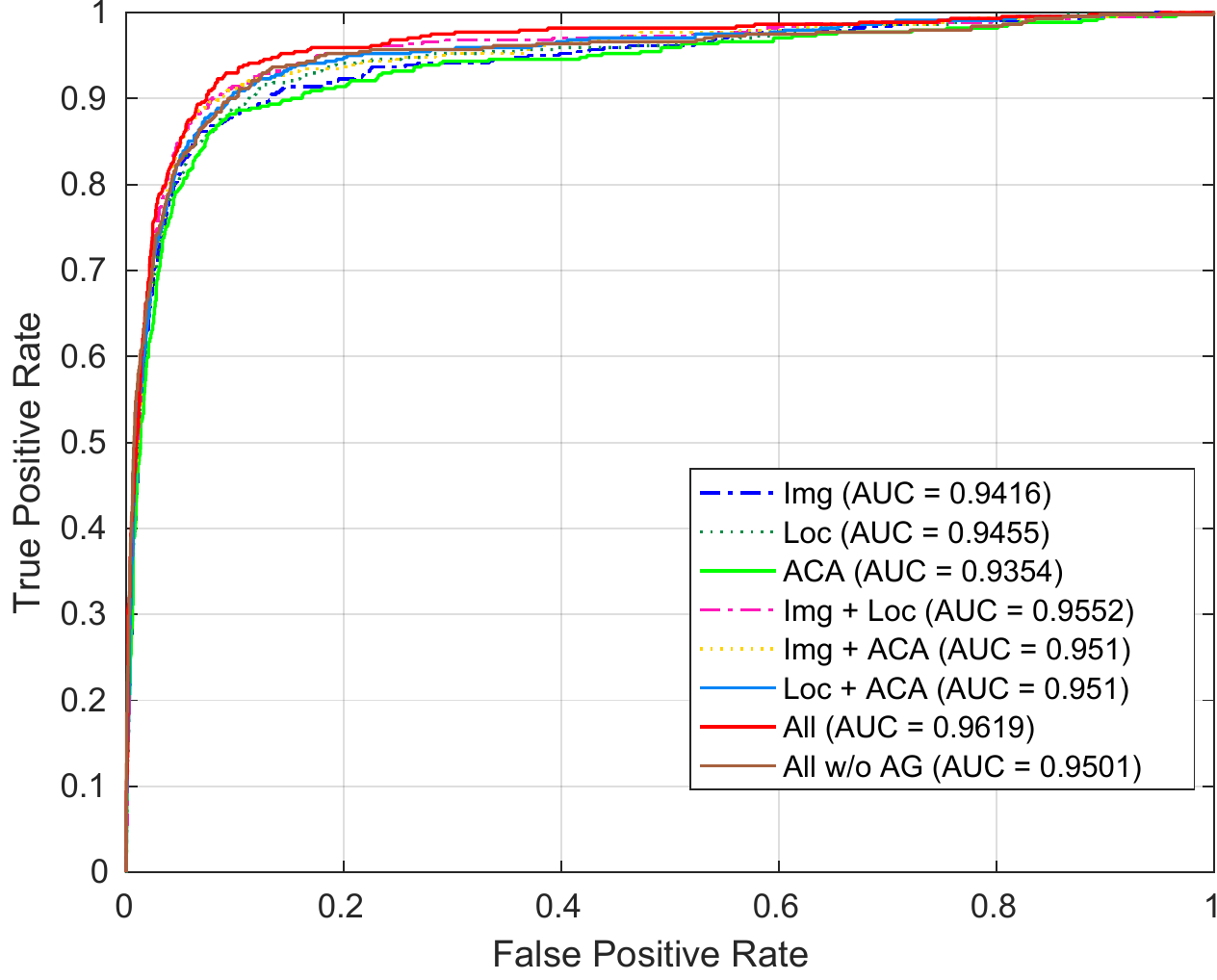}
        \label{exp_Visante_sub}}
    \subfigure[]
    {\includegraphics[width=.47\linewidth, height= .4\linewidth]{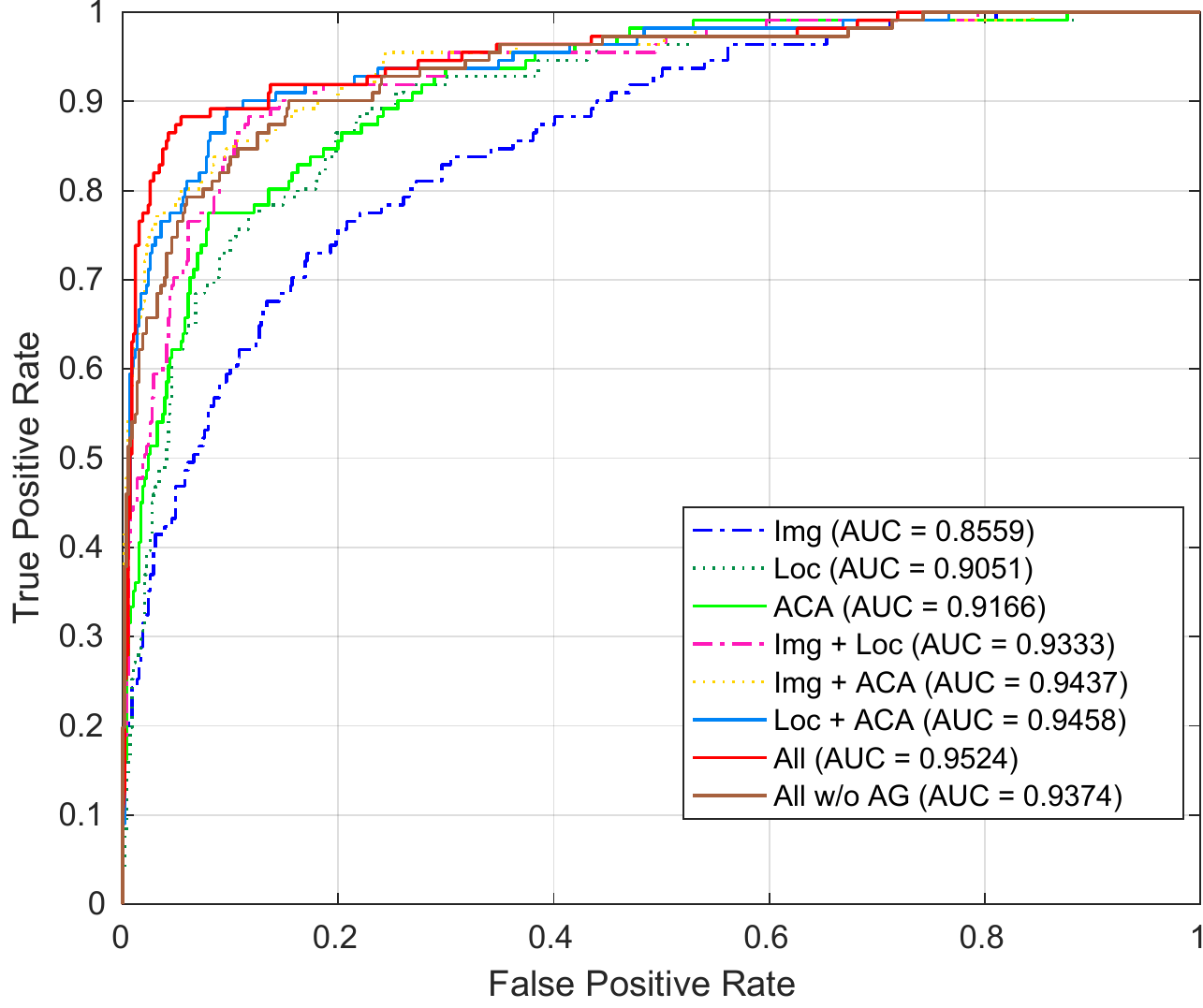}
        \label{exp_Cirrus_sub}}
    \caption {Performance of different sub-network combinations in our MLDN on the (A) Visante AS-OCT and (B) Cirrus HD-OCT datasets. }
    \label{exp_curve_sub}
\end{figure*}

Our multi-level architecture employs three sub-networks to generate the screening result. To evaluate the effectiveness of each sub-network, we report the results on different combinations of both datasets, as shown in Table~\ref{Tab_Sub_result} and Fig.~\ref{exp_curve_sub}. From the results, we can see that the performance on global images is the worst. Single sub-networks on local regions and ACA patches obtain similar scores. Combinations of two sub-networks perform better than a single sub-network, and our final architecture with all sub-networks obtains the best performance.
Moreover, we report results obtained without the intensity-based augmentation (Our MLDN w/o AG). We can observe that the intensity-based augmentation yields clear improvements, with the AUC score increasing from $0.950$ to $0.962$ on the Visante AS-OCT dataset and from $0.937$ to $0.952$ on the Cirrus HD-OCT dataset.

\section{Conclusion}
\label{sec_conclusion}

In this paper, we presented an automated angle-closure detection system for AS-OCT images. A multi-level deep network was designed to obtain multi-level AS-OCT representations on clinically meaningful regions. In addition, an intensity-based augmentation for AS-OCT was introduced in the training phase. Experiments on two clinical AS-OCT datasets demonstrate that the proposed MLDN method outperforms previous detection methods and other deep learning networks. The design of MLDN was guided by clinical prior knowledge, and the intensity-based augmentation could also be utilized in other OCT-related tasks. Moreover, our MLDN architecture is a general framework not limited to AS-OCT images. Actually, it can be used in any real-life application that would benefit from extraction of a meaningful foreground region and integrating it with global context. In the future, we will examine further applications of MLDN.

\ifCLASSOPTIONcaptionsoff
\newpage
\fi

{
    \bibliographystyle{IEEEtran}
    \bibliography{Deep_ASOCT}
}

\end{document}